# Cutting-Splicing data augmentation: A novel technology for medical image segmentation

Lianting Hu[1,2], PhD; Huiying Liang[1,2], PhD; Jiajie Tang[3,4], PhD; Xin Li[3], PhD; Li Huang[3], MS; Long Lu[3,4], PhD

1. *Medical Big Data Center, Guangdong Provincial People's Hospital/Guangdong Academy of Medical Sciences, Guangzhou, Guangdong 510080, China*
2. *Guangdong Cardiovascular Institute, Guangzhou, Guangdong 510080, China*
3. *School of Information Management, Wuhan University, Wuhan, China*
4. *Guangzhou Women and Children's Medical Center, Guangzhou Medical University, Guangzhou, China*

**Corresponding Author:** Long Lu, PhD. School of Information Management, Wuhan University, No 16, Luojiashan Road, Wuchang District, Wuhan, 430072, China. Phone: +86-18986022408.

Email: lulong@whu.edu.cn.

## Abstract

**Background:** Medical images are more difficult to acquire and annotate than natural images, which results in data augmentation technologies often being used in medical image segmentation tasks. Most data augmentation technologies used in medical segmentation were originally developed on natural images and do not take into account the characteristic that the overall layout of medical images is standard and fixed.

**Methods:** Based on the characteristics of medical images, we developed the cutting-splicing data augmentation (CS-DA) method, a novel data augmentation technology for medical image segmentation. CS-DA augments the dataset by splicing different position components cut from different original medical images into a new image. The characteristics of the medical image result in the new image having the same layout as and similar appearance to the original image. Compared with classical data augmentation technologies, CS-DA is simpler and more robust.

Moreover, CS-DA does not introduce any noise or fake information into the newly created image.

**Results:** To explore the properties of CS-DA, many experiments are conducted on eight diverse datasets. On the training dataset with the small sample size, CS-DA can effectively increase the performance of the segmentation model. When CS-DA is used together with classical data augmentation technologies, the performance of the segmentation model can be further improved and is much better than that of CS-DA and classical data augmentation separately. We also explored the influence of the number of components, the position of the cutting line, and the splicing method on the CS-DA performance.

**Conclusions:** The excellent performance of CS-DA in the experiment has confirmed the effectiveness of CS-DA, and provides a new data augmentation idea for the small sample segmentation task.

**Keywords:** cutting-splicing, data augmentation, medical image, segmentation, deep neural networks

## Introduction

Medical image segmentation can separate regions or objects of interest from other parts of the image (1). It is the first step of most analysis procedures (2). The segmentation result can provide a reliable basis for clinical diagnosis and pathology research, and assist the doctor in making a more accurate diagnosis (3). The victory of the deep neural network (DNN) in October 2012 ImageNet (4) marked the beginning of DNN's rapid development. Over the next few years, DNN achieved state-of-the-art performance in countless natural image classification (5), detection (6), and segmentation (7) tasks. The excellent performance of DNN resulted in it quickly being introduced into the field of medical image segmentation. DNN requires a large number of parameters to fit the mapping function between the input and the output (8). These parameters need to be optimized by a huge annotated dataset; otherwise, over fitting will appear (9). In medical image segmentation tasks, annotations are made by radiologists with expert knowledge on the data and task, and most annotations of medical images are time-consuming (10). Additionally, most medical images contain the patient's personal private information, so sharing medical images among multiple sites is not always possible. Therefore, it is difficult to collect a huge annotated dataset in medical image segmentation tasks. For DNN to achieve better performance in medical image segmentation, data augmentation (DA) is often used to increase the sample size of the medical image training dataset (11).

Classical data augmentation (Cla-DA) technologies are developed almost entirely based on natural images. After these Cla-DA technologies succeed in natural images, they are introduced into the field of medical images. Most of those introductions are successful. Technologies such as flipping (12, 13), cropping (12, 14), padding (15, 16), and rotation (17, 18) have been shown to improve the performance of segmentation models in medical image segmentation tasks. However, some introductions are infeasible, such as color augmentation (11) and color space transformation (19). Both of the two above DA technologies perform some operations on the three-color channels of the natural image, but most kinds of medical images only have a single-color channel. Therefore, color augmentation and color space transformation are rarely used in the field of medical image segmentation. The generative adversarial network (GAN) is currently the most popular DA technology and also was developed based on natural images (19). GANs have achieved good performance in the field of medical image classification (20, 21) and also been used in medical image segmentation as the segmentation framework (22). The GAN cannot generate a corresponding segmentation-object mask for the generated fake image, so it has not yet been applied to medical image segmentation as a DA technology. DA technologies successfully introduced from the field of natural images have achieved good results in medical image segmentation tasks. However, these DA technologies were not originally developed based on medical images, so some characteristics of medical images are not taken into account in these DA technologies.

There is a difference between medical images and natural images in terms of acquisition. The acquisition of natural images is extremely easy. A person who has not received any training can acquire several natural images in a few seconds using a smart phone without any complex preparation. However, the acquisition of medical images needs to meet many harsh conditions (23): sick patients, professional radiologists, expensive scanning equipment, and long acquisition times. The acquisition can only be done in the hospital, and the patient must pay a considerable fee for this. Obviously, the acquisition of medical images is more difficult than that of natural images, which is also one of the reasons why it is hard to collect a huge annotated dataset in the medical image field (24). The difficulty in acquiring medical images is a double-edged sword because it makes DNN segmentation models unable to be fully trained, but it also makes the overall layout of medical images more standard and fixed in a specific segmentation task. For example, in pulmonary X-ray images that are acquired in the axial plane, the right lung and the left lung always appear on the left side and the right side respectively (**Fig. 1**). However, the overall layout of natural images is more variable and random. For instance, the elephant's head can appear on the left side, the right side, or the middle of natural images, which will not appear on a fixed location (**Fig. 1**).

At present, none of the Cla-DA technologies applied in medical image segmentation have taken into account the characteristic that the overall layout of medical images is standard and fixed. In this study, based on the characteristic of medical image layout, we propose cutting-splicing data augmentation (CS-DA) that is a novel DA technology specifically for medical image segmentation. The idea of CS-DA is simple: original medical images are cut into multiple components, all in the same way; then different position components cut from different original images splice together to form a new image. Assume that the sample size of the original medical image dataset is five, and each original medical image is cut into four components. Thus 620 ($5^4$ – 5 = 620) new images can be created. Because of the characteristic of the medical image layout, the same position components cut from different original images have almost the same part of the object or the background. As a result, the new image spliced from those components has the same overall layout as the original image. Compared with Cla-DA, CS-DA is simpler and more robust. The original image does not need to be input into complex mathematical functions; it only needs to be concatenated in the matrix format. More importantly, Cla-DA technologies generate new images by randomly changing the information of original images, which introduces fake information into the new image. By contrast, CS-DA does not introduce any noise or fake information into the original image. All information in the new image created by CS-DA comes from the original image.

The rest of the paper is organized as follows: In the *Methods* section, we present the difference between medical images and natural images in six aspects and explain CS-DA in detail. Next, we conduct various experiments on eight datasets to explore the properties of CS-DA in the *Results* section. The experimental results are discussed in the Discussion section, and the paper concludes in the last section.

# Methods

## Differences Between Medical Images and Natural Images

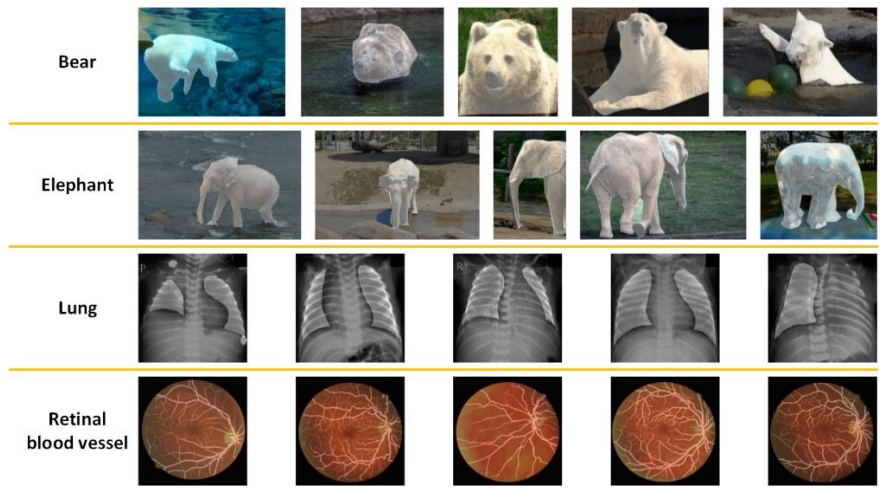

**Fig. 1.** Natural images of "bear" and "elephant." and medical images of "lung" and "retinal blood vessel." The brightness of the object region in each image has been enhanced. Images of bears and elephants are from the COCO dataset (25). The pulmonary X-ray images are acquired from Guangzhou Women and Children's Medical Center (26). The retinal blood vessel images are from the DRIVE dataset (27).

There are some key differences between natural images and medical images. These differences are reflected in six aspects: camera orientation, object posture, object location, object completeness, object scale, and object category.

*1) Camera orientation*

In a natural scene, there is no specific requirement for the orientation of the camera. The object can be photographed from any orientation. As shown in **Fig. 1**, the camera can be located opposite the bear or under the bear. However, in the clinic, the orientation of the scanner or other image acquisition equipment is fixed. For example, the pulmonary X-ray images are all scanned in the posteroanterior view (**Fig. 1**).

*2) Object posture*

The object in the natural scene can be in any posture during the photographing process. For example, in **Fig. 1**, the bear can play ball or swim. However, a patient must hold a specific posture during the scanning process. For younger children who cannot control themselves, radiologists will even use auxiliary devices or sedation to immobilize their bodies. Therefore, the scanned organ also maintains a specific shape on the medical image.

*3) Object location*

In a natural scene, the photographer will arrange the position of the object in the natural image according to his or her layout ideas. There is no fixed pattern in the layout design of the natural image, so the position of the object is always random. In the clinic, the radiologist will adjust the scanner to ensure that the organ is at the center of the image, or a specific location.

*4) Object completeness*

Natural scenes are complex and changeable, and sometimes there are obstructions between the object and the camera that prevent the object from being completely displayed in the natural image. For example, part of the bear's body in **Fig. 1** is blocked by the water in the pond. Medical images are acquired in a much cleaner scene. Unusual obstructions are not allowed between the patient and the scanner, so the integrity of the object can be guaranteed.

*5) Object scale*

The pixel size information is not recorded in the natural image, so the real size of the object cannot be calculated by the object region in the natural image. On the contrary, the scale is the basic information presented in the medical image. For example, the pixel spacing and the slice thickness is typically provided in the file header. That information can help us to normalize objects in different images into a standard space.

*6) Object category*

In a segmentation task of natural images, categories of objects are too numerous to be finely defined. Therefore, the object is assigned a coarse category. In **Fig. 1**, "bear" is a coarse category, which can be defined finely as "brown bear" or "polar bear." On the contrary, human organs have been clearly defined. Each segmented region in the medical image has a clear category.

Natural images have many possibilities in the above six aspects. Uncertainty in these six aspects makes natural images diverse. Therefore, there are huge differences among natural images containing the same kind of object. On the other hand, the diversity of medical images is constrained in the six aspects by the standard scanning equipment, well-trained radiologists, and patients who strictly follow the scan specifications. Consequently, in a specific segmentation task, the overall layout of medical images is standard and fixed. This characteristic makes the overall layout among medical images consistent.

## Cutting and Splicing Images

The consistency of the overall layout results in interchangeability among regions in the same position of different medical images. A new medical image can be created by using a region in one medical image to replace the same position region in another medical image. The new medical image has the same overall layout as the original medical image, and the object in the new medical image is intact. In a segmentation task, the new medical image can be mixed with original medical images to train the segmentation model.

Based on the region interchangeability of medical images, the CS-DA is proposed in this study. The technology includes two steps: cutting images into components, and splicing components into new images.

*1) Cutting images into components*

In this step, the original image is cut into multiple components. Each cutting line is perpendicular to a certain dimension of the image and crosses through the entire image. The *n* cutting lines in the same dimension divide the image into (*n*+1) equal components. In 2D images, there are two dimensions that can be cut (**Fig. 2a**), and 3D images have three dimensions that can be cut (**Fig. 2b**). All images in the same dataset are cut in the same way. Therefore, the number of components cut out from each image is the same. The same way of the image is also applied to its segmentation-object mask.

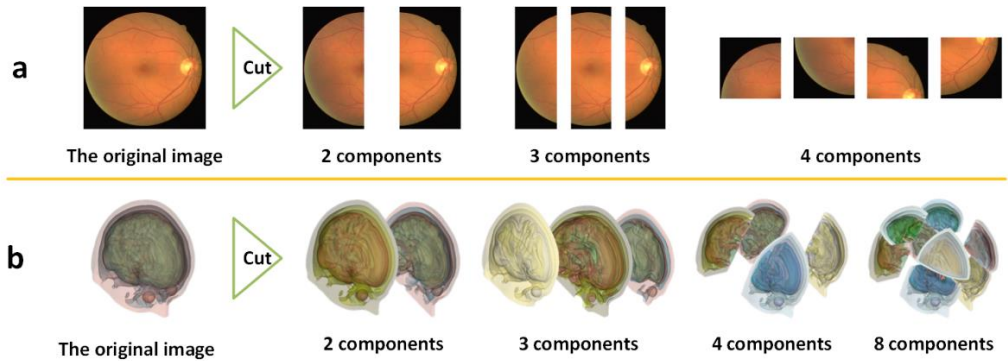

**Fig. 2.** Cutting images into components. **a**, cutting the 2D retinal blood vessel image into two, three, and four components respectively. **b**, cutting the 3D brain image into two, three, four, and eight components respectively.

*2) Splicing components into new images*

This step can be performed by two methods: normal splicing (NorS), and symmetrical splicing (SymS). In NorS, there are large differences among components cut from the same one image, so there is no interchangeability among those components. The component of the original image in a specific position only can be used to fill the same position region in the new image (**Fig. 3a**). The entire structure of the human body is almost bilaterally symmetrical. What's more, some organs inside the human body are also almost bilaterally symmetrical, such as the lung. If the medical image is perpendicular to the symmetrical plane of the symmetrical organ, the medical image is also symmetrical. In the symmetrical image, if a cutting line is on the symmetry axis, the components on both sides of the cutting line are similar. Therefore, there is interchangeability between the component on the one side and the component that has been flipped on the other side. In this kind of case, SymS can be performed. The flipped component of the original image can be used to fill the symmetry position region of the new image (**Fig. 3b**). In this step, mask components corresponding to the image components spliced into a new image will also be spliced into a new mask. The new mask is the segmentation gold standard for the new image.

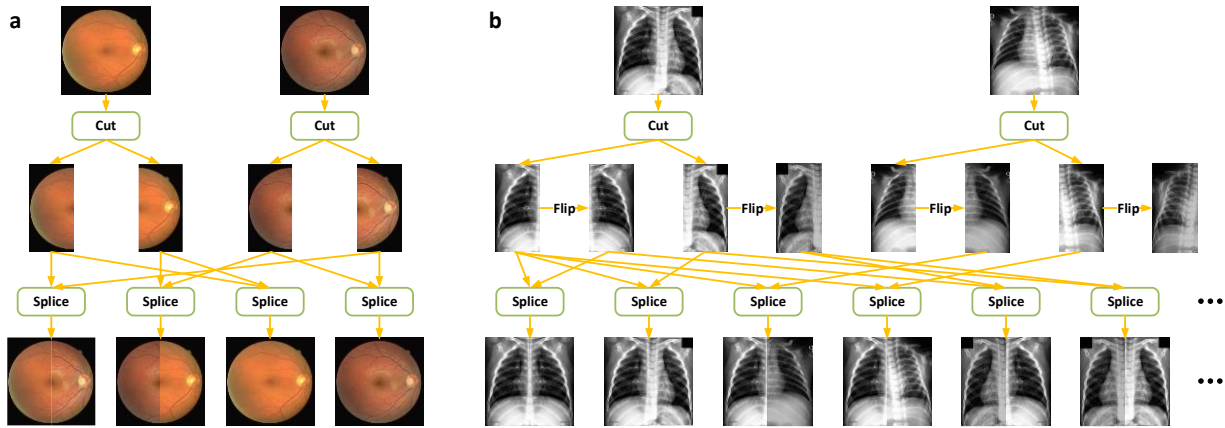

**Fig. 3.** Splicing components into new images. **a**, creating new retinal blood vessel images by NorS. **b**, creating new pulmonary X-ray images by SymS. In **a** and **b**, the sample size of the original image dataset is two, and one original image is cut into two components.

Some natural images and medical images created by CS-DA are shown in **Fig. 4**. The diversity of the overall layout in the natural image makes the new image abnormal. The completeness of the object in new natural images is broken. By contrast, the new medical image looks normal. Every new medical image has a complete object.

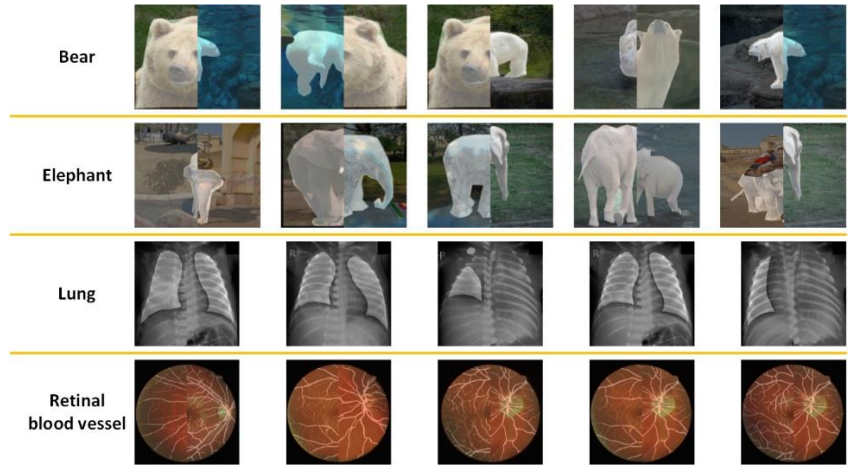

**Fig. 4.** Newly created natural images and medical images. All images are created by cutting-splicing two images from the same dataset as **Fig. 1**. The brightness of the object region in each image has been enhanced.

## Histogram Matching

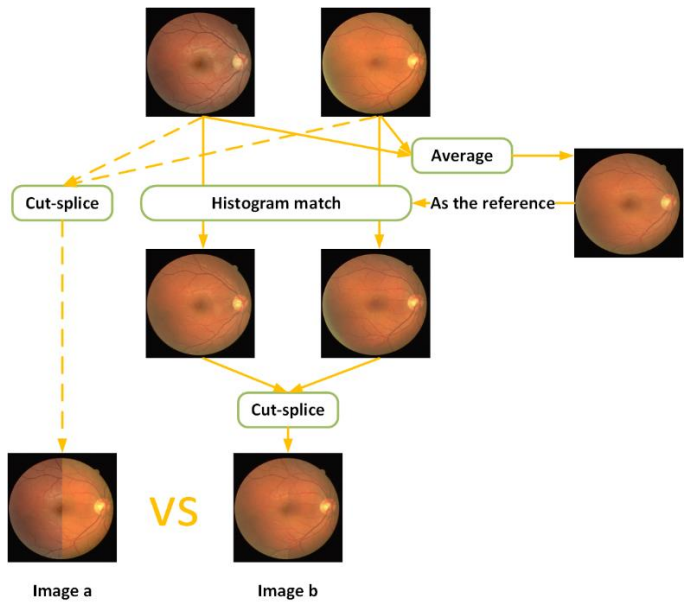

**Fig. 5.** Comparing the effects of using and not using histogram matching in CS-DA. Dotted line arrows and solid line arrows represent the process without histogram matching and with histogram matching respectively. Image **a** and image **b** are newly created images.

The components in a new image are from multiple original images with different styles. As a result, the style of the new image is mixed. However, the style of the real image that the trained segmentation model will be used to segment is simple. Therefore, it is necessary to normalize the styles of all components in a new image.

In the study, histogram matching (28) is used to normalize styles. Given the source image and the reference image, let $r$ and $z$ be the pixel value of the source image and the reference image respectively. $p_r(r)$ and $p_z(z)$ are used to represent the continuous probability density functions (PDFs) of $r$ and $z$. $p_r(r)$ and $p_z(z)$ can be obtained by the following equations:

$$p_r(r_k) = \frac{n_{r_k}}{N_r}, (k = 0,1,2, \cdots, L-1) \tag{1}$$

, and

$$p_z(z_k) = \frac{n_{z_k}}{N_z}, (k = 0,1,2, \cdots, L-1) \tag{2}$$

where $n_{r_k}$ and $n_{z_k}$ are the number of pixels with the value $k$ in the source image and the reference image respectively. $N_r$ and $N_z$ are the total number of pixels for the source image and the reference image. The range of pixel value is $[0, L-1]$. Let $s$ be the random variable with the following peculiarity in probability:

$$s = T(r) = (L-1) \int_0^r p_r(w)dw \tag{3}$$

where $w$ is a dummy variable of integration. The right side of the equation is recognized as the cumulative distribution function of the random variable $r$ (28). Next, let's define another random variable $z$ with the following probability:

$$G(z) = (L-1) \int_0^z p_z(t)dt = s \tag{4}$$

where $t$ is a dummy variable of integration. Based on Eq. 3 and Eq. 4, $G(z) = T(r)$ can be gotten. Therefore, $z$ must satisfy the following condition:

$$z = G^{-1}(s) = G^{-1}[T(r)] \tag{5}$$

After the above calculation, $z$ is the pixel value of the image converted from the source image by histogram matching.

A single original image in the training dataset is special, and its style may not cover as many test images as possible. Therefore, in the study, the average image of all images in the dataset, rather than a specific image, is selected as the reference image in histogram matching. In the study, histogram matching is performed twice. The first time, all images in the training dataset are histogram-matched to their average image. Here, histogram matching works the same as histogram equalization to decrease the differences among all training images. The second time, before cutting-splicing, all original images that are used to create the same new image are histogram-matched to their average image (**Fig. 5**). This time is to decrease the differences among components in the new image. As shown in **Fig. 5**, image **a** is created without histogram matching. On the contrary, image **b** is created with histogram matching. As a result, the style of image **b** is much simpler than image **a**.

## Sample Size of Augmented Dataset

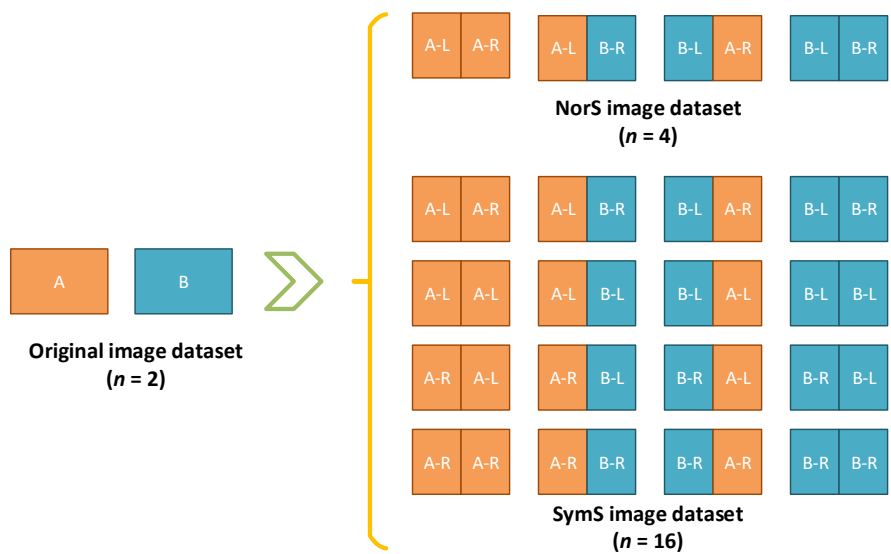

**Fig. 6.** The example of the two splicing methods to create new images. Each new image is spliced by two components. "*-L" and "*-R" represent the left component and the right component of the original image "*" respectively. $n$ represents the sample size of the dataset.

CS-DA as a DA technology, the sample size of the augmented dataset is determined by the sample size of the original dataset, the number of components in the new image, and the splicing method. **Fig. 6** is an example of the two splicing methods to create new images. Next, the sample size of the augmented dataset will be introduced based on the example. In the original dataset, there are two original images: A, and B. Each original image is cut into two components: the left component and the right component. For convenience of explanation, we define some symbols: "*-L" and "*-R" represent the left component and the right component of the original image "*" respectively; "+" represents splicing two components together.

If the original image is asymmetrical, NorS is used to create new images. Components in the new image can come from different original images, such as A-L+B-R, and can also come from the same original image, such as A-L+A-R. The new image A-L+A-R actually is the original image A. Therefore, the original dataset is a subset of the augmented dataset. The sample size is increased from two in the original dataset to four in the augmented dataset. SymS will be used to create new images if the original image is symmetrical. The left component in the new image can be the left component of the original image, such as B-L+A-R, and can also be the flipped right component of the original image, such as B-R+A-R. SymS doubles the choices of the component in the new image. Therefore, the sample size of the SymS dataset is four times that of the NorS dataset.

Based on the above example, we can find that new images can be regarded as permutations of components from original images, and repetition is allowed in the permutation. Therefore, a more general calculation method of the augmented dataset sample size can be summarized as follows. $N$ indicates the sample size of the original dataset. $k$ indicates the number of components cut from an original image. If the original image is asymmetrical, the sample size of the augmented dataset can be calculated by the following equation:

$$N_n = N^k \qquad (6)$$

where $N_n$ indicates the sample size of the augmented dataset created by NorS. If the original image is symmetrical, the sample size can be calculated by the following equation:

$$N_s = (2N)^k \qquad (7)$$

where $N_s$ indicates the sample size of the augmented dataset created by SymS.

## Results

## Datasets

In this study, multiple segmentation datasets are prepared to comprehensively evaluate the performance of CS-DA. These datasets contain multiple modalities, multiple segmentation-objects, multiple scanning planes, and multiple dimensions (**Table 1**).

The pediatric chest radiograph (PCR) dataset is an X-ray image dataset acquired at the Guangzhou Women and Children's Medical Center (26). There are 852 normal X-ray images and 499 abnormal X-ray images with different pathological conditions in the PCR dataset. The lung mask is annotated by a radiologist with three years of clinical experience. To guarantee accurate annotating, a junior radiologist with 10 years of clinical experience checked and calibrated the annotated masks (26).

The Coronavirus Disease 2019 CT segmentation (COVID-19-CT) dataset contains 100 axial CT images from about 40 patients with Coronavirus Disease 2019 that were converted from openly accessible JPG images found in Italian Society of Medical and Interventional Radiology. This dataset is created by two radiologists: Tomas Sakinis, and Håvard Bjørke Jenssen (https://www.medseg.ai/). The lung masks are contributed by Johannes Hofmanninger (29).

**Table 1.** Datasets

| Dataset | Modality | Objects | Dimensions | Original sample size | Image size |
|---------|----------|---------|------------|----------------------|------------|
| PCR | X-ray | lung (coronal plane) | 2D | 1351 | 320×320 |
| COVID-19-CT | CT | lung (axial plane) | 2D | 100 | 512×512 |
| DRIVE | fundus image | retinal blood vessel | 2D | 20 | 576×560 |
| CHAOS-T1in-2D | MRI | liver, spleen, right kidney, left kidney | 2D | 80 | 224×144 |
| CHAOS-T1out-2D | MRI | liver, spleen, right kidney, left kidney | 2D | 80 | 224×144 |
| CHAOS-T2-2D | MRI | liver, spleen, right kidney, left kidney | 2D | 80 | 224×144 |
| CHAOS-T2-3D | MRI | liver, spleen, right kidney, left kidney | 3D | 20 | 224×144×24 |
| IBSR | MRI | GM, WM, CSF | 3D | 20 | 80×96×80 |

The Digital Retinal Images for Vessel Extraction (DRIVE) dataset has been established to enable comparative studies on the segmentation of blood vessels in retinal images (https://drive.grand-challenge.org/). The DRIVE dataset is acquired from a diabetic retinopathy screening program in the Netherlands. A total of 40 images in the DRIVE dataset are equally divided into a training dataset and a test dataset. For the training image, a single manual segmentation of the vasculature is available (27). In the study, only the training dataset is used to train and test our approach.

The Combined Healthy Abdominal Organ Segmentation (CHAOS) challenge aims at the segmentation of abdominal organs (liver, kidneys, and spleen) from CT and MRI data (30). CHAOS challenge provides training data and corresponding organ masks of 20 different patients, but the organ masks of the test data are not available. To train and test our approach locally, only the training dataset is used in the study. We choose the MRI data to segment abdominal organs. The training dataset of CHAOS contains three MRI sequences: T1-DUAL in phase, T1-DUAL out phase, and T2-SPIR, each of which is being routinely performed to scan the abdomen using different radiofrequency pulse and gradient combinations. From each MRI sequence of each patient, four 2D slices in the axial plane are selected. Each selected slice contains all segmentation-objects (liver, spleen, right kidney, and left kidney). Therefore, a total of 80 2D slices can be selected from each MRI sequence. Those selected 2D slices formed three datasets: CHAOS T1-DUAL in phase 2D (CHAOS-T1in-2D) dataset, CHAOS T1-DUAL out phase 2D (CHAOS-T1out-2D) dataset, and CHAOS T2-SPIR 2D (CHAOS-T2-2D) dataset. Before training the segmentation framework, the three datasets need to be divided into training datasets and test datasets. Slices of the same patient will be divided into the same dataset. The original CHAOS T2-SPIR (CHAOS-T2-3D) dataset is directly used as a 3D dataset without selecting slices.

The Internet Brain Segmentation Repository (IBSR) provides manually guided expert segmentation results along with magnetic resonance brain image data. The 20 normal T1-weighted MR brain data datasets and their manual segmentations were provided by the Center for Morphometric Analysis at Massachusetts General Hospital and are available at http://www.cma.mgh.harvard.edu/ibsr/. The segmentation-objects of the dataset are gray matter (GM), white matter (WM), cerebrospinal fluid (CSF).

The segmentation framework in the study is the U-Net (31) that requires the input image with some specific size. Therefore, all 2D datasets and the CHAOS-T2-3D dataset are scaled and cropped to the required image sizes that are closest to their original image sizes. The IBSR dataset is normalized to the required image size by SPM (32).

## Experimental Setup

All datasets in **Table 1** are divided into the test dataset and the basic training dataset. 20% of the samples in the original dataset are randomly selected to form the test dataset. To evaluate the effects of DA technologies under different sample sizes of the training dataset, multiple basic training datasets are formed by randomly selecting samples from the remaining 80% of the original dataset. For instance, in the DRIVE dataset, the sample size of the test dataset is 4, and the sample sizes of basic training datasets are 3, 5, 8, 10, and 15.

Based on one basic training dataset, three augmented training datasets are created by different DA technologies (**Fig. 7**). The first augmented training dataset is created by Cla-DA technologies that contain Gaussian blur, cropping or padding, piecewise affine, and affine transformations. For the 3D image, those Cla-DA technologies are applied sequentially to the 2D slices. Therefore, all slices at the axial plane get the same augmentation parameters. Each original image produces 10 new images. The second augmented training dataset is created by CS-DA. The third augmented training dataset is created by both Cla-DA and CS-DA. CS-DA is applied to the basic training dataset that has been augmented by Cla-DA technologies. The configuration of Cla-DA and CS-DA is the same as that of the first two augmented training datasets.

The number of new images created by CS-DA is so large that it costs too much memory space to store those new images. For example, given a basic training dataset of the sample size of 200, the sample size will increase to 40,000 ($200^2$) after only being augmented by CS-DA, while the sample size will increase to 4,840,000 (($200\times11)^2$) after being augmented by both Cla-DA and CS-DA. To avoid storing such a large augmented training dataset, we do not create all new images all at once. Before using the augmented dataset, the original image index of the component in new images is obtained by listing all permutations of all original images. In an iteration, only a small batch of images will be inputted into the DNN segmentation framework. Therefore, we only create the required images by their original image indexes. The created image will be overwritten in the next iteration.

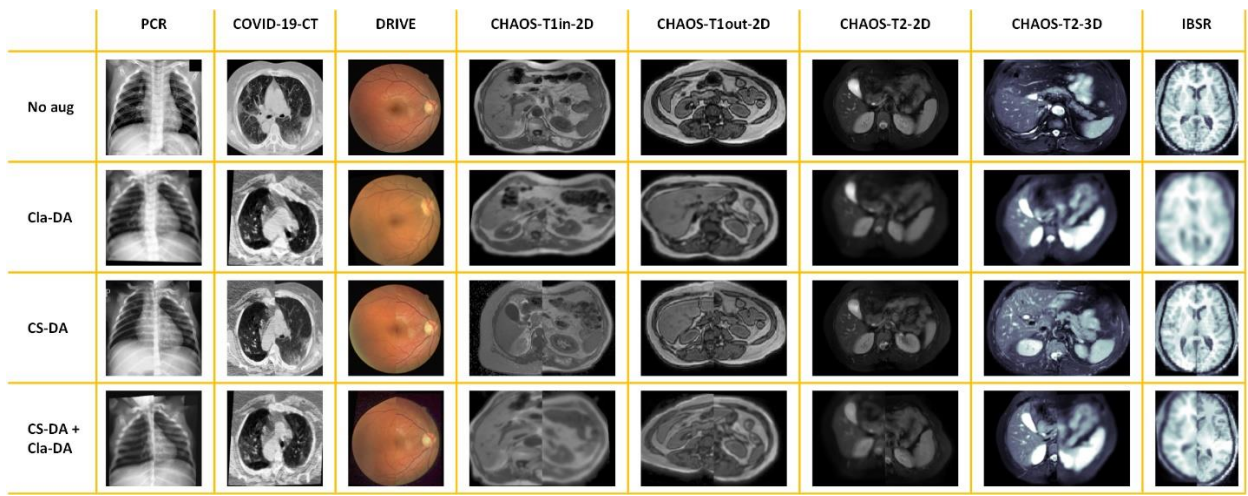

**Fig. 7.** Examples of the basic and augmented training datasets. Only one slice in the axial plane is shown in 3D images. The number of components in CS-DA is 2.

The U-Net is the segmentation framework in this study. For 2D images, the original U-Net (31) is used. Apart from the padding method of all convolutional layers are replaced from "valid" to "same," there are no other modifications in the framework. For 3D images, the 3D version U-Net (33) is used. The framework configuration is consistent with the original paper without any modification. The loss function is generalized dice loss (34), and the learning rate is 0.0001. When the overall error of the training data no longer increases, the model stops training. The average dice similarity coefficient (DSC) (35)s is used as the accuracy measure to compare the ground truth and the predicted object volume. The average DSC is calculated by taking the average of the DSC of each class.

As mentioned above, the test and the basic training dataset are randomly selected from the original dataset. What's more, the sample sizes of some basic training datasets are very small. Under the same model configuration, the results obtained by running the model multiple times may be unstable. In order to make the results stable and comparable, the mean result of the model running three times is treated as the final evaluation measure. Meanwhile, for a fairer comparison of different DA methods, all DA technologies are applied to the same three basic training datasets when the model is run three times. The samples of multiple basic training datasets are gradually increased. In order to make the results of multiple basic training datasets comparable, the basic training dataset with the smaller sample size is required to be a subset of the larger basic training dataset.

## Model Comparison

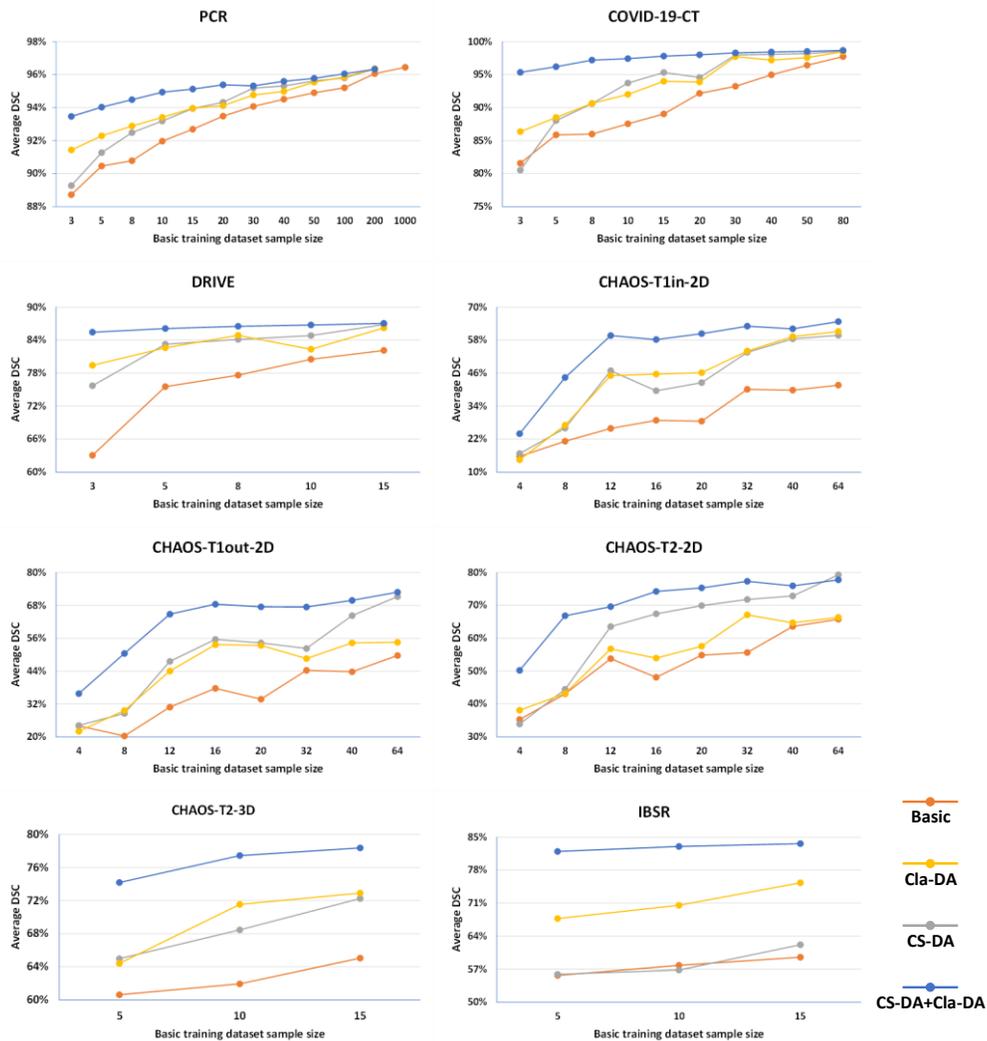

**Fig. 8.** The test average DSC for different DA technologies. In PCR dataset, when the sample size of the basic training dataset is 1000, the sample size of the augmented dataset can be $1 \times 10^8$ which is difficult to learn. Therefore, DA technologies are not applied to the basic training

dataset with the sample size of 1000.

In this section, the effect of different DA technologies on small sample datasets is explored. In CS-DA, all images are cut into 2 components, and the cutting line is in the first dimension. Only NorS is used to splice new images. **Fig. 8** presents the test average DSC for different DA technologies. The application of CS-DA technologies improves the segmentation performance of U-Net on the basic training dataset. The improvement gradually decreases with the increase in the sample size of the basic training dataset. CS-DA and Cla-DA have similar improvements in segmentation results. It is difficult to draw a unified conclusion about which is better from the results in these eight datasets. The combination of CS-DA and Cla-DA has the greatest effect on segmentation results. In all eight datasets, the combination of these two methods has achieved the best results.

## Number of Components

In this section, the effect of the number of components on the segmentation performance is explored. 2D images are cut into two, three, and four components. 3D images are cut into two, three, four, and eight components. The ways of cutting are consistent with those of 2D images and 3D images in **Fig. 2**. Only NorS is used in this section. In another group of experiments, Cla-DA technologies are also applied to augment the training dataset (**Fig. 9**). Without Cla-DA, in addition to the IBSR dataset, the test average DSC increases with the increase in the number of components. When the number of components is sufficient, the performance without Cla-DA will be close to or exceed that with Cla-DA. But in the IBSR dataset, the test average DSC is not much improved with the increase in the number of components. When CS-DA is used together with Cla-DA, after the number of components exceeds two, the test average DSC increases slightly, or stays steady, or even decreases.

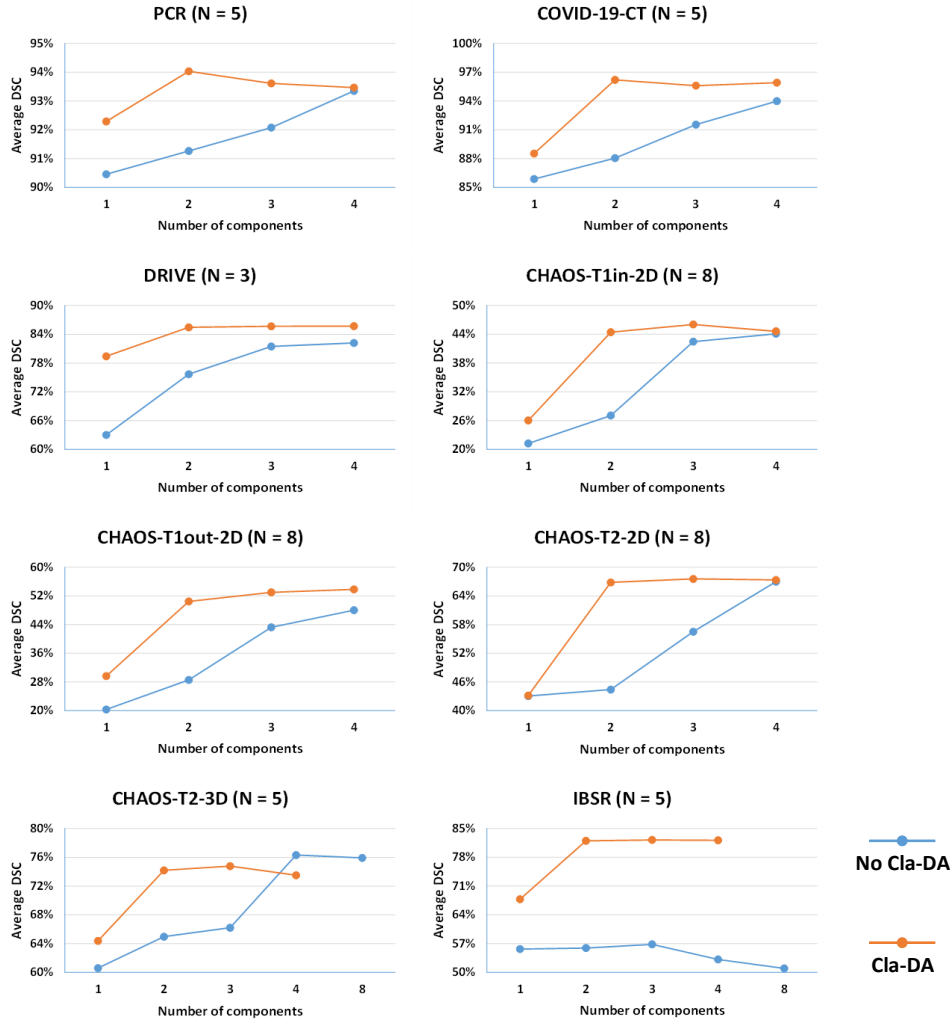

**Fig. 9.** The test average DSC on different numbers of components. The "N" in the title of each subgraph indicates the sample size of the basic training dataset. In the CHAOS-T2-3D dataset and the IBSR dataset, when the number of components is eight, the sample size of the augmented training dataset that is augmented by CS-DA and Cla-DA is $55^8$ which is difficult to learn. Therefore, this experiment is not conducted.

## Position of the Cutting Line

There are multiple choices for the position of each cutting line. In the 2D image, a cutting line can be located in the first dimension or the second dimension. Meanwhile, the 3D image has three dimensions to choose from. Therefore, a question needs to be explored: is the position of the cutting line related to the effect of CS-DA? In this section, experiments are conducted to answer the question. CS-DA is applied to the basic training dataset with the small sample size. The original image is cut into two components by one cutting line. The cutting line is in the first dimension, or the second dimension, or the third dimension (only for 3D images). The segmentation model trials more than 10 times on the augmented training dataset and the average DSC is collected. The DSC population in each cutting configuration has unequal variances and unequal sample sizes. Therefore, we use Welch's t-test (36) to test the hypothesis that two populations have equal means. The results of the experiment are shown in **Table 2**. All *P*-values in **Table 2** are greater than 0.05. This means that we can accept the null hypothesis that the two populations are not significantly different with a 95% confidence. In other words, the position of the cutting line has no significant effect on CS-DA.

**Table 2.** The test DSC on different cutting line positions

| Dataset | Sample size of basic training dataset | 1st dim DSC | 2nd dim DSC | 3rd dim DSC | *P*-value |
|---------|---------------------------------------|-------------|-------------|-------------|-----------|
| PCR | 5 | 92.46% | 92.62% | - | 0.207 |
| COVID-19-CT | 5 | 88.34% | 88.18% | - | 0.771 |
| DRIVE | 3 | 74.78% | 74.50% | - | 0.732 |
| CHAOS-T1in-2D | 8 | 37.50% | 38.57% | - | 0.116 |
| CHAOS-T1out-2D | 8 | 40.11% | 38.63% | - | 0.283 |
| CHAOS-T2-2D | 8 | 58.88% | 57.84% | - | 0.458 |
| IBSR | 5 | 55.20% | 54.12% | | 0.541 |
| IBSR | 5 | 55.20% | | 54.75% | 0.750 |
| IBSR | 5 | - | 54.12% | 54.75% | 0.733 |
| CHAOS-T2-3D | 5 | 75.46% | 74.82% | - | 0.166 |
| CHAOS-T2-3D | 5 | 75.46% | - | 74.77% | 0.100 |
| CHAOS-T2-3D | 5 | - | 74.82% | 74.77% | 0.935 |

## Splicing Method on Symmetrical Images

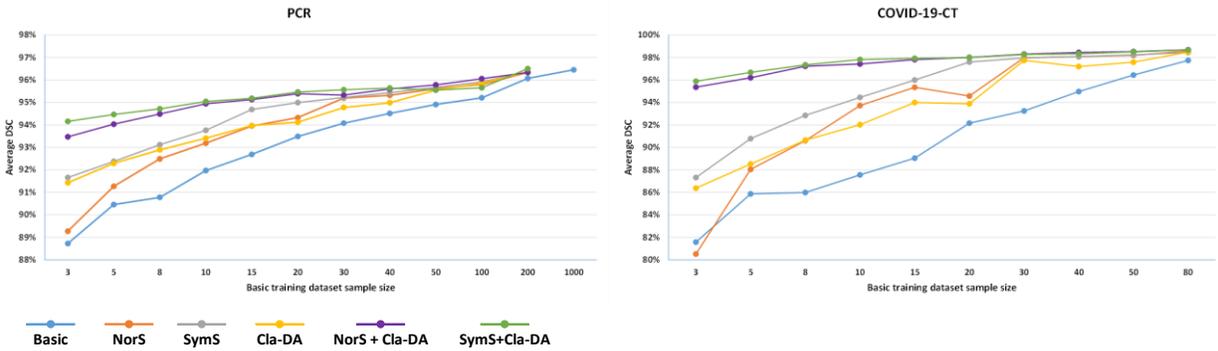

**Fig. 10.** The test dataset average DSC in datasets of symmetrical images.

In symmetrical images, both NorS and SymS can be used to splice new images. According to Eq. 6 and Eq. 7, the number of new images augmented by SymS is more than that by NorS. In this section, the performance of these two splicing methods is explored. The lungs are almost bilaterally symmetrical in the coronal plane and the axial plane. Therefore, SymS can be used in the PCR and COVID-19-CT datasets. In experiments, the number of the components is two, and the cutting line is in the first dimension. **Fig. 10** shows the test dataset DSC on symmetrical datasets. Compared with the performance of NorS on multiple basic training datasets, the performance of SymS is much better. When the basic training dataset has fewer samples, the performance advantage of SymS is more obvious. As mentioned in the *Model Comparison* section, compared to Cla-DA technology, CS-DA with NorS has no obvious advantages. However, from the results shown in **Fig. 10**, we can see that CS-DA with SymS outperforms the Cla-DA technology. When used with Cla-DA technologies, the advantages of SymS over NorS are not so obvious.

## Discussion

In this study, CS-DA successfully augmented the medical image dataset of multiple modalities (MRI, CT, and X-ray), multiple organs (lungs, abdominal organs, eyes, and brain), multiple scanning planes (coronal and axial planes) and multiple dimensions (2D and 3D). Control experiments comprehensively confirmed that CS-DA greatly improved the performance of medical image segmentation tasks that have small sample size. Meanwhile, those experiments provide some suggestions for the application of CS-DA in practice. As shown in **Fig. 9**, the more components, the more improvement CS-DA can bring to the segmentation model. This is mainly caused by the fact that the more components, the more augmented images (Eq. 6 and Eq. 7). However, too many images in the augmented training dataset is not always a good thing, as this situation requires too much time to be learned by the segmentation model. Therefore, choosing the appropriate number of components can weigh the segmentation performance against training efficiency. It can be found from **Table 2** that the position of cutting lines has no significant effects on the segmentation performance. This discovery makes the application of CS-DA easier. We do not need to worry that the cutting line destroys the integrity of the segmentation-object, and can design the image cutting scheme more flexibly. As shown in **Fig. 10**, the effect of SymS is better than those of NorS and Cla-DA. Therefore, if the medical image is symmetrical, SymS, rather than NorS, should be selected as the splicing method.

CS-DA is a novel DA technology for medical image segmentation tasks and can achieve a similar performance as Cla-DA. Moreover, in symmetrical images, the effect of CS-DA even exceeds that of Cla-DA. The goal of proposing CS-DA is not to replace Cla-DA technologies, but to provide a new way of thinking for augmenting images. The Cla-DA creates a large number of new images by the randomization of parameters. For example, the width of cropping or padding, the standard deviation of the Gaussian blur, and the position of piecewise affine are all randomly set. A new image is defined by the permutation of those random set parameters. The randomization of parameters makes new images different, but also uncontrollable. Meanwhile, all Cla-DA technologies will change the pixel values and the layout of the images. Those changes introduce fake information into the original image and make new images that cannot exist in the real world. By contrast, CS-DA creates a large number of new images by permutations of components from original images. The only parameter that needs to be set in CS-DA is the number of components. What's more, the detailed process only contains cutting and splicing. There is no huge computation in CS-DA, and the image does not require transformation by complex mathematical functions. The new image is uniquely determined. More importantly, CS-DA will not introduce any fake information into the original images and can maintain the layout and the object scale of the original image. All information in the CS-DA created image is from original images. Surprisingly, there is no conflict between CS-DA and Cla-DA. They can work together on the same dataset. Moreover, when the number of components is appropriate, the combined use of CS-DA and Cla-DA can further improve the effectiveness of the segmentation model.

It can be inferred from Eq. 6 and Eq. 7 that the more components we divide, the more augmented images we will obtain. More new images can adequately train the segmentation model. Therefore, having a higher number of components has a positive effect on the performance of the segmentation model. However, medical images are just similar, rather than identical, to each other. If the medical image is cut into too many components, the similarity among components in the same position of different original images will decrease. As a result, the similarity of layouts between the new image and the test image will also decrease. Therefore, having more components also has a negative effect on the performance of the segmentation model. As shown in **Fig. 9**, when only CS-DA is used, the test average DSC increases with the increase of the number of components, which suggests that the positive effect is greater than the negative effect in this case. When CS-DA is used together with Cla-DA, after the number of components is greater than two, the test average DSC increases only slightly, or stays steady, or even decreases. This reason is that Cla-DA reduces the similarity of the layout among images. As a result, the negative effect of having more components is enhanced.

In CS-DA, two questions remain to be discussed: 1) Does the cutting line that passes through the segmentation-object have effects on the performance of the segmentation model? 2) Does the unevenness between components have effects on the performance of the segmentation model? In the PCR dataset, the cutting line in the first dimension lies between the left lung and the right lung, which maintains the integrity of the lung. However, the cutting line in the second dimension passes through the lung and cuts the lung into the upper part and the lower part. As shown in **Table 2**, there is no significant difference between the test average DSC of the two cutting line cases, which indicates whether the cutting line passes through the segmentation-object does not affect the segmentation performance. For the second question, although histogram matching is used to try to unify the style of original images, there is still a big difference among the distribution of components from different original images. This makes the pixel values in the connection edge between components abrupt and uneven. We have tried to use Gaussian smoothing in the connection edge to reduce the unevenness. The result suggests that Gaussian smoothing cannot affect the performance, but destroys the original information at the connection edge. At present, there is no useful method to eliminate the unevenness. However, according to the augmentation result of CS-DA, the adverse effects brought by such unevenness do not offset the beneficial effects brought by the increase in sample size.

CS-DA can be further improved. Future work can focus on the following two aspects. 1) There is no limit to the number of components, which makes the augmented training dataset too large when the number of components is relatively large. Such a large dataset needs to be learned by the more complex segmentation framework, and also requires more powerful hardware equipment in order to execute the complex segmentation framework. 2) The interchangeability among components cut from different original images is the key factor for CS-DA to work. In clinical practice, the interchangeability may not be perfect. Therefore, before cutting and splicing, performing medical image normalization/registration and segmentation-object detection may improve the interchangeability.

## Conclusions and Future Work

In this study, we pointed out that the current DA technologies applied in medical image segmentation have not taken into account the characteristic that the overall layout of medical images is standard and fixed in a specific task. The main contribution of our work is proposing the CS-DA, a novel DA technology for medical image segmentation based on the above characteristic. In CS-DA, histogram matching is used to eliminate the difference in style between components cut from different original images. The sample size of the augmented training dataset can be adjusted by the number of components and the splicing method. Through multiple experiments on eight different datasets, the properties of CS-DA were explored. We discussed the application prospects of CS-DA in practice and the difference between CS-DA and Cla-DA. At the same time, the influence of the number of components, the splicing method, the position of the cutting line, and the unevenness of the component edge on the performance of the segmentation model was analyzed.

## Acknowledgements

Lianting Hu designed the study and wrote the original draft. Huiying Liang and Long Lu reviewed the manuscript. Xin Li and Li Huang preprocessed some data. This work was supported by the National Natural Science Foundation of China [grant numbers 61772375, 18ZDA325]; Hubei Provincial Natural Science Foundation of China [grant number 2019CFA025]; Independent Research Project of School of Information Management Wuhan University [grant number 413100032]. The numerical calculations in this paper have been done on the supercomputing system in the Supercomputing Center of Wuhan University. The authors would like to thank the anonymous reviewers for their valuable suggestions.